\documentclass[a4paper,12pt]{article} \usepackage{hyperref}
\usepackage{amssymb,amsmath,color,times}

\newcommand{\BEAS}{\begin{eqnarray*}}
\newcommand{\EEAS}{\end{eqnarray*}}
\newcommand{\BEA}{\begin{eqnarray}}
\newcommand{\EEA}{\end{eqnarray}}
\newcommand{\BEQ}{\begin{equation}}
\newcommand{\EEQ}{\end{equation}}
\newcommand{\BIT}{\begin{itemize}}
\newcommand{\EIT}{\end{itemize}}
\newcommand{\BNUM}{\begin{enumerate}}
\newcommand{\ENUM}{\end{enumerate}}
\newcommand{\BA}{\begin{array}}
\newcommand{\EA}{\end{array}}

\newcommand{\tr}{\mathop{ \rm tr}}

\newcommand{\idm}{I}
\newcommand{\rb}{\mathbb{R}}
\newcommand{\BlackBox}{\rule{1.5ex}{1.5ex}}  

\newenvironment{proof}{\par\noindent{\bf Proof\ }}{\hfill\BlackBox\\[2mm]}

\newtheorem{proposition}{Proposition}

\newcommand{\mysec}[1]{Section~\ref{sec:#1}}
\newcommand{\eq}[1]{Eq.~(\ref{eq:#1})}

\author{ Francis Bach, Julien Mairal, Jean Ponce \\
Willow Project-team \\
Laboratoire d'Informatique de l'Ecole Normale Sup\'erieure \\
(CNRS/ENS/INRIA UMR 8548)\\
45, rue dÕUlm, 75230 Paris, France}
\title{ Convex Sparse Matrix Factorizations }

\begin{document}
\maketitle

\abstract{We present a convex formulation of dictionary learning for sparse signal decomposition. Convexity is obtained by replacing the usual explicit upper bound on the dictionary size  by a convex rank-reducing term similar to the trace norm. In particular, our formulation introduces an explicit trade-off between size and sparsity of the decomposition of rectangular matrices. Using a large set of synthetic examples, we compare the estimation abilities of the convex and non-convex approaches, showing that while the convex formulation has a single local minimum, this may lead in some cases to performance which is inferior to the local minima of the non-convex formulation.

}

\section{Introduction}

Sparse decompositions have become prominent tools in signal processing~\cite{chen99atomic}, image processing~\cite{elad}, machine learning, and statistics~\cite{lasso}.
Many relaxations and approximations of the associated minimum cardinality problems are now available, based on greedy approaches~\cite{matchingpursuit} or convex relaxations through the $\ell^1$-norm~\cite{chen99atomic,lasso}.
Active areas of research are  the design of efficient algorithms to solve the optimization problems associated with the convex non differentiable norms (see, e.g.,~\cite{lars}), the  theoretical study of the sparsifying effect of these norms~\cite{Cand05,Zhaoyu}, and the learning of the dictionary directly from data (see, e.g.,~\cite{olshausen,elad}).

In this paper, we focus on the third problem---namely, we assume that we are given a matrix  $Y \in \rb^{N \times P}$ and we look for factorizations of the form $X = UV^\top$, where $U \in \rb^{N \times M}$ and $V \in \rb^{P \times M}$, that are close to $Y$ and such that the matrix $U$ is sparse. This corresponds to decomposing $N$ vectors in $\rb^P$ (the rows of $Y$) over a dictionary of size $M$. The columns of $V$ are the \emph{dictionary elements} (of dimension $P$), while the rows of $U$ are the \emph{decomposition coefficients} of each data point. Learning sparse dictionaries from data has shown great promise in signal processing tasks, such as image or speech processing~\cite{elad}, and core machine learning tasks such as clustering may be seen as special cases of this framework~\cite{ding}.

Various approaches have been designed for sparse dictionary learning.
Most of them  consider a specific loss between entries of $X$ and $Y$, and directly optimize over $U$ and $V$, with additional constraints on $U$ and $V$~\cite{ksvd,ng}: dictionary elements, i.e., columns of $V$, may or may not be constrained to unit norms, while a penalization on the rows of $U$ is added to impose sparsity. Various forms of jointly non-convex alternating optimization frameworks may then be used~\cite{ksvd,ng,elad}. The main goal of this paper is to study the possibility and efficiency of convexifying these non-convex approaches. As with all convexifications, this leads to the absence of non-global local minima, and should allow simpler analysis. However, does it really work in the dictionary learning context? That is, does convexity lead to better decompositions?

While in the context of sparse decomposition with \emph{fixed} dictionaries, convexification has led to both theoretical and practical improvements~\cite{Cand05,lasso,Zhaoyu}, we report both positive and negative results  in the context of dictionary learning. That is, convexification sometimes helps and sometimes does not. In particular, in high-sparsity and low-dictionary-size situations, the non-convex fomulation outperforms the convex one, while in other situations, the convex formulation does perform  better (see \mysec{simulations} for more details). 
 
The paper is organized as follows:  we show in \mysec{decomposition} that if  the size of the dictionary is not bounded, then dictionary learning may be naturally  cast as a convex optimization problem; moreover, in \mysec{closed}, we show that in many cases of interest, this problem may be solved in closed form, shedding some light on what is exactly achieved and not achieved by these formulations. Finally, in \mysec{mixed}, we propose a   mixed $\ell^1$-$\ell^2$ formulation that leads to both low-rank and sparse solutions in a joint convex framework. In \mysec{simulations}, we present simulations on a large set of synthetic examples.

\textbf{Notations} \hspace*{.2cm}  Given a rectangular matrix $X \in \rb^{N \times P}$ and
$n \in \{1,\dots,N\}, p \in \{1,\dots,P\}$, we   denote by $X(n,p)$ or $X_{np}$ its element indexed by the pair $(n,p)$, by $X(:,p) \in \rb^N$ its $p$-th column and by $X(n,:) \in \rb^P$ its $n$-th row.
Moreover, given a vector $x \in \rb^N$, we denote by $\| x\|_q$ its $\ell^q$-norm, i.e., for $q \in [1,\infty)$, $\|x\|_q =  ( \sum_{n=1}^N | x_n |^q  )^{1/q}$ and $\| x\|_\infty = \max_{ n \in \{1,\dots,N \}} |x_n|$.
We also  write a matrix $U \in \rb^{ N \times P}$ as $U =[ u_1, \dots, u_M ]$, where each $u_m \in \rb^N$.

\section{Decomposition norms}

\label{sec:decomposition}

We consider a loss $\ell:\rb\times \rb \to \rb$ which is convex with respect to the second variable. We assume in this paper that all entries of $Y$ are observed and   the risk of the estimate $X$ is equal to$ \frac{1}{NP}\sum_{n=1}^N \sum_{p=1}^P \ell(Y_{np},X_{np})$. Note that our framework extends in a straightforward way to matrix completion settings by summing only over observed entries~\cite{Srebro2005Maximum}.

We consider factorizations of the form $X = UV^\top$; 
in order to constrain $U$ and $V$, we consider the following
optimization problem:
\BEQ
\label{eq:UV}
\min_{U \in \rb^{N \times M} ,V \in \rb^{ P \times M} } \frac{1}{NP}\sum_{n=1}^N \sum_{p=1}^P \ell(Y_{np},(UV^\top)_{np})
+ \frac{\lambda}{2} \sum_{m=1}^M ( \| u_m \|_{C}^2 + \| v_m \|_{R}^2),
\EEQ
where $\| \cdot \|_{C}$ and $\| \cdot \|_{R}$ are any \emph{norms} on $\rb^N$ and $\rb^P$ (on the \textbf{c}olumn space and \textbf{r}ow space of the original matrix $X$). This corresponds to penalizing each column of $U$ and $V$.
In this paper, instead of considering $U$ and $V$ separately, we consider the matrix $X$ and the set of its decompositions on the form $X=UV^\top$, and in particular, the one with minimum sum of norms $ \| u_m \|_{C}^2$, $  \| v_m \|_{R}^2$, $m \in \{1,\dots, M\}$.
That is, for $X \in \rb^{N\times P}$, we consider
\BEQ
\label{eq:defF}
    f_{D}^M(X)=  \min_{
 (U,V) \in 
\rb^{N \times M} \times \rb^{P \times M} , \
  X = UV^\top
} \ \
\frac{1}{2}\sum_{m=1}^M 
 ( \| u_m \|_{C}^2 + \| v_m \|_{R}^2). \EEQ
 If $M$ is strictly smaller than the rank of $X$, then we let $f_D^M(X)=+\infty$. Note that the minimum is always attained if $M$ is larger than or equal to the rank of $X$.
 Given $X$, each pair $(u_m,v_m)$ is defined up to a scaling factor, i.e.,
 $(u_m,v_m)$ may be replaced by $(u_m s_m,v_m s_m^{-1})$; optimizing with respect to $s_m$  leads to the following equivalent formulation:
 \BEQ
 \label{eq:UVfromX}
  f_{D}^M(X)
 =     \min_{
 (U,V) \in 
\rb^{N \times M} \times \rb^{P \times M} , \
  X = UV^\top
}
\sum_{m=1}^M 
\| u_m \|_{C}  \| v_m \|_{R}.
\EEQ
Moreover, we may derive another equivalent formulation by constraining the norms of the columns of $V$ to one, i.e.,
\BEQ
 \label{eq:UVfromXcons}
  f_{D}^M(X)
 =     \min_{
 (U,V) \in 
\rb^{N \times M} \times \rb^{P \times M} , \
  X = UV^\top, \ \forall m, \|v_m\|_R=1
} \ \ 
\sum_{m=1}^M 
\| u_m \|_{C}  .
\EEQ
This implies that constraining dictionary elements to be of unit norm, which is a common assumption in this context~\cite{ng,elad}, is equivalent to penalizing the norms of the decomposition coefficients instead of the squared norms.

Our optimization problem  in \eq{UV} may now be equivalently written as
\BEQ
\label{eq:UVX}
\min_{X \in \rb^{N \times P} } \frac{1}{NP} \sum_{n=1}^N \sum_{p=1}^P \ell(Y_{np},X_{np})
+ \lambda f_{D}^M(X).
\EEQ
with any of the three formulations of $f_D^M(X)$ in Eqs.~(\ref{eq:defF})-(\ref{eq:UVfromXcons}). 
The next proposition shows that if the size $M$ of the dictionary is allowed to grow, then  we obtain a norm on rectangular matrices, which we refer to as a \emph{decomposition} norm. In particular, this shows that if $M$ is large enough the problem in \eq{UVX} is a convex optimization problem.

\vspace*{-.1cm}

\begin{proposition} For all $X \in \rb^{N \times P}$, the limit 
$f^\infty_{D}(X)=\lim_{M \to \infty} f^M_{D}(X)$ exists and $f^\infty_{D}(\cdot)$ is a norm on rectangular matrices.
\end{proposition}

\vspace*{-.2cm}

\begin{proof}
Since given $X$, $ f^M_{D}(X)$ is nonnegative and clearly nonincreasing with $M$, it has a nonnegative limit when $M$ tends to infinity.
The only non trivial part is the triangular inequality, i.e.,
$f^\infty_{D}(X_1+X_2 )\leqslant f^\infty_{D}(X_1) + f^\infty_{D}(X_2)$. Let $\varepsilon>0$ and let
$(U_1,V_1)$ and $(U_2,V_2)$ be the two $\varepsilon$-optimal decompositions, i.e., such that $ 
f^\infty_{D}(X_1  ) \geqslant \sum_{m=1}^{M_1} 
\| u_{1m} \|_{C}  \| v_{1m} \|_{R} - \varepsilon  
$
 and  $ 
f^\infty_{D}(X_2  ) \geqslant \sum_{m=1}^{M_2} 
\| u_{2m} \|_{C}  \| v_{2m} \|_{R} - \varepsilon  
$.  Without loss of generality, we may asssume that $M_1=M_2=M$. We consider $U = [ U_1 \ U_2 ]$,
$V = [ V_1 \ V_2 ]$, we have $X = X_1 + X_2 = UV^\top$ and
$f^\infty_{D}(X  ) \leqslant \sum_{m=1}^M ( 
\| u_{1m} \|_{C}  \| v_{1m} \|_{R}  
+
\| u_{2m} \|_{C}  \| v_{2m} \|_{R}
)   \leqslant  f^\infty_{D}(X_1) + f^\infty_{D}(X_2) + 2\varepsilon $. We obtain the triangular inequality by letting $\varepsilon$ tend to zero.
\end{proof}
Following the last proposition, we now let $M$ tend to infinity; that is, if we denote $\| X\|_{D} = f^\infty_{D}(X)$, we consider the following rank-unconstrained and \emph{convex} problem:
\BEQ
\label{eq:UVXinfinity}
\min_{X \in \rb^{N \times P} } \frac{1}{NP} \sum_{n=1}^N \sum_{p=1}^P \ell(Y_{np},X_{np})
+ \| X\|_D.
\EEQ
However, there are three potentially major caveats that should be kept in mind: 

\textbf{Convexity and polynomial time} \hspace*{.2cm}
 Even though the norm $\| \cdot \|_{D}$ leads to a convex function, computing or approximating it may   take exponential time---in general, it is not because a problem is convex that it can be solved in polynomial time. In some cases, however, it may be computed in closed form, as presented in \mysec{closed}, while in other cases, an efficiently computable convex lower-bound is available (see \mysec{mixed}).

\textbf{Rank and dictionary size} \hspace*{.2cm}
 The dictionary size $M$ must be allowed to grow to obtain convexity and there is no reason, \emph{in general}, to have a finite $M$ such that
$f^\infty_D(X)=f^M_D(X)$. In some cases presented in \mysec{closed}, the optimal $M$ is finite, but we conjecture that in general the required $M$ may be unbounded. Moreover, in non sparse situations, the  rank of $X$ and the dictionary size $M$ are usually equal, i.e., the matrices  $U$ and $V$ have full rank. However, in sparse decompositions, $M$ may be larger than the rank of $X$, and sometimes even larger than the underlying data dimension $P$ (the corresponding dictionaries are said the be {\em overcomplete}). 

\textbf{Local minima} \hspace*{.2cm}
 The minimization problem in \eq{UV},  with respect to $U$ and $V$, even with $M$ very large, may still have multiple local minima, as opposed to the one in $X$, i.e., in \eq{UVXinfinity}, which has a single local minimum. The main reason is that the optimization problem defining $(U,V)$ from $X$, i.e., \eq{UVfromX}, may itself have multiple local minima.
In particular, it is to be constrasted to the optimization problem
\BEQ
\label{eq:UVconvex}   
\min_{U \in \rb^{N\times M},V^{N \times M}} \frac{1}{NP} \sum_{n=1}^N \sum_{p=1}^P \ell(Y_{np},(UV^\top)_{np})
+  {\lambda} \| U V^\top \|_{D},
\EEQ
which will turn out to have no local minima if $M$ is large enough  (see \mysec{lowrank} for more details).

Before looking at special cases, we compute the dual norm of $\| \cdot \|_{D}$  (see, e.g.,~\cite{boyd} for the definition and properties of dual norms), which will be used later.

\vspace*{-.1cm}

\begin{proposition}[Dual norm]
\label{prop:dualnorm}
The dual norm $\| Y\|_{D}^\ast$, defined as
$$
\| Y\|_{D}^\ast = \sup_{ \| X\|_{D} \leqslant 1} \tr X^\top Y
,$$ is equal to
$
\| Y\|_{D}^\ast = \sup_{\|u\|_{C} \leqslant 1, \ \|v\|_{R} \leqslant 1 } v^\top Y^\top u
$.
\end{proposition}

\vspace*{-.2cm}

\begin{proof}
We have, by convex duality (see, e.g.,~\cite{boyd}),

\vspace*{-.5cm}

\BEAS
\| Y\|_{D}^\ast  & = &    \sup_{ \| X\|_{D} \leqslant 1} \tr X^\top Y
= \inf_{ \lambda \geqslant 0} \sup_{ X } \tr X^\top Y - \lambda  \| X\|_{D} + \lambda \\[-.1cm]
& = &    \inf_{ \lambda \geqslant 0} \lim_{M \to \infty}\sum_{m=1}^M (\sup_{ u_m, v_m } v_m^\top Y^\top u_m
 - \lambda  \|u_m\|_{C} \|v_m\|_{R} )+ \lambda
\EEAS

\vspace*{-.3cm}

Let $a = \sup_{\|u\|_{C} \leqslant 1, \ \|v\|_{R} \leqslant 1 } v^\top Y^\top u$. If $\lambda < a$, $$\sup_{ u_m, v_m } v_m^\top Y^\top u_m
 - \lambda  \|u_m\|_{C} \|v_m\|_{R} = +\infty, $$while if $\lambda > a$, then
 $$\sup_{ u_m, v_m } v_m^\top Y^\top u_m
 - \lambda  \|u_m\|_{C} \|v_m\|_{R} = 0.$$ The result follows.
\end{proof}

 \vspace*{-.200cm}
 
\section{Closed-form decomposition norms}
\label{sec:closed}

We now consider important special cases, where the decomposition norms can be expressed in closed form. For these norms, with the square loss, the convex optimization problems may also be solved in closed form.
Essentially, in this section, we show that in simple situations involving sparsity (in particular when one of the two norms $\|\cdot \|_C$ or $\| \cdot \|_R$ is the $\ell^1$-norm), letting the dictionary size $M$ go to infinity often leads to trivial dictionary solutions, namely a copy of some of the rows of $Y$. This shows the importance of constraining not only the $\ell^1$-norms, but also the $\ell^2$-norms, of the sparse vectors $u_m$, $m \in \{1,\dots,M\}$, and leads to the joint low-rank/high-sparsity solution presented in \mysec{mixed}.

\subsection{ Trace norm: $ \| \cdot \|_{C} =  \| \cdot \|_2   $ and  $ \| \cdot \|_{R} =  \| \cdot \|_2   $ }

\label{sec:tracenorm}
When we constrain both the $\ell^2$-norms of $u_m$ and of $v_m$,
it is well-known, that $ \| \cdot \|_{D}$ is   the sum of the singular values of $X$, also known as the trace norm~\cite{Srebro2005Maximum}. In this case we only need $M \leqslant \min \{N,P\}$ dictionary elements, but this number will turn out in general to be a lot smaller---see in particular\cite{tracenorm} for rank consistency results related to the trace norm. Moreover, with the square loss, the solution of the optimization problem
in \eq{UVX} is  $X = \sum_{m=1}^{\min \{N,P\}} \max \{\sigma_m - \lambda NP,0\}  u_m v_m^\top$, where $Y = 
 \sum_{m=1}^{\min \{N,P\}} \sigma_m u_m v_m^\top$ is the singular value decomposition of $Y$. Thresholding of singular values, as well as its interpretation as trace norm minimization is well-known and well-studied. However, sparse decompositions (as opposed to simply low-rank decompositions) have shown to lead to better decompositions in many domains such as image processing (see, e.g., \cite{olshausen}).

\subsection{ Sum of norms of rows: $ \| \cdot \|_{C} =  \| \cdot \|_1   $  }
\label{sec:rows}

When we use the $\ell^1$-norm for $\|u_m\|_C$, whatever the norm on $v_m$, we have:
\BEAS
\| Y\|_{D}^\ast &  = &     \sup_{\|u\|_1 \leqslant 1, \ \|v\|_{R} \leqslant 1 } v^\top Y^\top u
=   \sup_{  \|v\|_{R} \leqslant 1 } \sup_{\|u\|_1 \leqslant 1} v^\top Y^\top u 
=   \sup_{  \|v\|_{R} \leqslant 1 }\| Y v\|_\infty \\[-.1cm]
& = &     \max_{n \in \{1,\dots,N \} } \max_v \| Y(n,:) v \|_{R}
= \max_{n \in \{1,\dots,N \} } \| Y(n,:)^\top \|_{R}^\ast,
\EEAS
which implies immediately that 
$$   
\| X\|_{D}\! =\! \sup_{ \|Y\|_{D}^\ast \leqslant 1} \tr X^\top Y\! = \!  \sum_{n=1}^N 
 \sup_{ \|Y(n,:)^\top\|_{R}^\ast \leqslant 1} \tr X(n,:)Y(n,:)^\top  \! = \!  \sum_{n=1}^N \| X(n,:)^\top \|_{R}.
$$
That is, the decomposition norm is simply the sum of the norms of the rows.   Moreover, an optimal decomposition is
$X = \sum_{n=1}^N \delta_n \delta_n^\top X$, where $\delta_n \in \rb^N$ is a vector with all null components except at $n$, where it is equal to one. In this case, each row of $X$ is a dictionary element and the decomposition is indeed extremely sparse (only one non zero coefficient).

In particular, when $\| \cdot \|_{R} = \| \cdot \|_2$, we obtain the sum of the $\ell^2$-norms of the rows, which leads to a closed form solution to \eq{UVXinfinity} as $X(n,:) =  \max\{ \|Y(n,:)^\top\|_2 - \lambda NP , 0 \} Y(n,:)/\|Y(n,:)^\top\|_2  $ for all $n \in \{1,\dots,N\}$.   
Also, when $\| \cdot \|_{R} = \| \cdot \|_1$, we obtain the sum of the $\ell^1$-norms of the rows, i.e, the $\ell^1$-norm of all entries of the matrix, which leads to decoupled equations for each entry   and closed form solution $X(n,p) =   \max\{ |Y(n,p)|  - \lambda NP , 0 \} 
 Y(n,p) / {| Y(n,p)| } $.   

These examples show that with the $\ell^1$-norm on the decomposition coefficients,  these simple decomposition norms do not lead to solutions with small dictionary sizes. This suggests to  consider a larger set of norms which leads to low-rank/small-dictionary \emph{and} sparse solutions. However, those two extreme cases still have a utility as they lead to good search ranges for the regularization parameter $\lambda$ for the mized norms presented in the next section.

\section{  Sparse decomposition norms  }
\label{sec:mixed}

We now assume that we have $\| \cdot \|_R = \| \cdot \|_2$, i.e, we use the $\ell^2$-norm on the dictionary elements. In this situation, when $\| \cdot \|_C = \| \cdot \|_1$, as shown in \mysec{rows}, the solution corresponds to a very sparse but large (i.e., large dictionary size $M$) matrix $U$; on the contrary, 
when $\| \cdot \|_C = \| \cdot \|_2$, as shown in \mysec{tracenorm}, we get a small but non sparse matrix $U$. It is thus natural to combine the two norms on the decomposition coefficients. The main result of this section is that the way we combine them is mostly irrelevant and we can choose the combination which is the easiest to optimize.

\vspace*{-.1cm}

\begin{proposition}
If the loss $\ell$ is differentiable, then for any function $f: \rb_+ \times \rb_+ \to \rb_+$, such that $\|\cdot \|_C = f( \| \cdot \|_1, \| \cdot \|_2)$ is a norm, and
which is increasing with respect to both variables, the solution of \eq{UVXinfinity} for  $\|\cdot \|_C = f( \| \cdot \|_1, \| \cdot \|_2)$ is the solution of \eq{UVXinfinity} for 
 $\|\cdot \|_C = [(1-\nu)\| \cdot \|_1^2 +  \nu\| \cdot \|_2^2]^{1/2}$, for a certain $\nu$ and a potentially different regularization parameter $\lambda$.
\end{proposition}

\vspace*{-.2cm}

\begin{proof} 
If we denote $L(X) = \frac{1}{NP}\sum_{n=1}^N\sum_{p=1}^P \ell(Y_{np},X_{np})$ and $L^\ast$ its Fenchel conjugate~\cite{boyd}, then the dual problem of \eq{UVXinfinity} is the problem of maximizing $-L^\ast(Y)$ such that $\|Y\|^\ast_D \leqslant \lambda$. Since the loss $L$ is differentiable, the primal solution $X$ is entirely characterized by the dual solution $Y$. The optimality condition for the dual problem is exactly that the gradient of $L^\ast$ is  equal to $uv^\top$, where $(u,v)$ is one of the maximizers in the definition of the dual norm, i.e., in $\sup_{ f( \| u \|_1, \| u \|_2) \leqslant 1, \ \|v\|_{2} \leqslant 1 } v^\top Y^\top u$. In this case, we have $v$ in closed form, and $u$ is the maximizer of 
$\sup_{ f( \| u \|_1, \| u \|_2) \leqslant 1 } u^\top Y Y^\top u$. With our assumptions on $f$, these maximizers are the same as the ones subject to  $\|u\|_1 \leqslant \alpha_1$ and $\|u\|_2 \leqslant \alpha_2$ for certain $\alpha_1$, $\alpha_2 \in \rb_+$. The optimality condition is thus independent of $f$. We then select the function $f(a,b) = 
 [(1-\nu)a^2 +  \nu b^2]^{1/2}$ which is  practical as it leads to simple lower bounds (see below).
\end{proof}

\vspace*{-.3cm}

We thus now consider the   norm   defined as $\|u \|_C^2 = (1-\nu)\| u \|_1^2 +  \nu\| u \|_2^2$. We denote by $F$ the convex function defined on symmetric matrices as  $
F(A) =  (1-\nu)   \sum_{i,j=1}^N |A_{ij}| +    \nu \tr A
$,
for which we have $F(uu^\top) = (1-\nu) \| u\|_1^2 + \nu \|u\|_2^2 = \| u\|_C^2$.

In the definition of $f_D^M(X)$ in \eq{defF}, we  can optimize with respect to $V$ in closed form, i.e.,
$$    \min_{ 
V \in  \rb^{P \times M}, \   X = UV^\top }
\frac{1}{2}\sum_{m=1}^M   \| v_m \|_2^2 = \frac{1}{2} \tr X^\top ( UU^\top)^{-1} X
 $$
 is
 attained at $V = X^\top  ( UU^\top)^{-1} U $ (the value is infinite if the span of the columns of $U$ is not included in the span of the columns of $X$). Thus the norm is equal to
 \BEQ
 \label{eq:N}   
 \| X \|_{D} =  \lim_{M \to \infty}
 \min_{ 
U \in  \rb^{N \times M}  }  \frac{1}{2} \sum_{m=1}^M F(u_m u_m^\top) + \frac{1}{2} \tr X^\top ( UU^\top)^{-1} X.
 \EEQ
Though $\| X \|_{D} $ is a convex function of $X$, we currently don't have a polynomial time algorithm to compute it, but, since 
$F$ is convex and homogeneous, 
 $\sum_{m\geqslant 0} F(u_m u_m^\top) \geqslant F(\sum_{m \geqslant 0} u_m u_m^\top)$. This leads to the following lower-bounding convex optimization problem in the positive semi-definite matrix $A = UU^\top$:
 \BEQ
 \label{eq:L}
     \| X \|_{D}  \geqslant \min_{A \in \rb^{N \times N}, \ A \succcurlyeq 0 }\frac{1}{2} F(A) + \frac{1}{2} \tr X^\top A^{-1} X.
\EEQ
 This  problem can now  be solved in polynomial time~\cite{boyd}.
This computable lower bound in \eq{L} may serve two purposes: (a) it provides a good initialization to gradient descent or path following rounding techniques presented in \mysec{rounding}; (b) the convex lower bound provides sufficient conditions for approximate \emph{global} optimality of the non convex problems~\cite{boyd}.

\subsection{Recovering the dictionary and/or the decomposition}
\label{sec:rounding}

Given a solution or approximate solution $X$ to our problem, one may want to recover dictionary elements $U$ and/or the decomposition $V$ for further analysis. Note that   (a) having one of them automatically gives the other one and (b) in some situations, e.g., denoising of $Y$ through estimating $X$, the matrices $U$ and $V$ are not explicitly needed.

We propose to iteratively minimize  with respect to $U$ (by gradient descent) the following function, which is a convex combination of the true function in \eq{N} and its upper bound in \eq{L}:
$$   \frac{1-\eta}{2} F(UU^\top) + \frac{\eta}{2}  \sum_{m\geqslant 0} F(u_m u_m^\top)  + \frac{1}{2} \tr X^\top ( UU^\top)^{-1} X.$$
When $\eta=0$ this is exactly our convex lower bound applied defined in \eq{L}, for which there are no local minima in $U$, although it is not a convex function of $U$ (see \mysec{lowrank} for more details), while at $\eta=1$, we get a non-convex function of   $U$, with potentially multiple local minima.  This path following strategy has shown to lead to good local minima in other settings~\cite{GNC}. 

Moreover, this procedure may be seen as the classical rounding operation that follows a convex relaxation---the only difference here is that we relax a hard convex problem into a simple convex problem. Finally, the same technique  can be applied when minimizing the regularized estimation problem in \eq{UVXinfinity}, and, as shown in \mysec{simulations}, rounding leads to better performance.



 

\subsection{Optimization with square loss   }

In our simulations, we will focus on the square loss as it leads to simpler optimization, but our decomposition norm framework could be applied to other losses.
With the square loss,  we can optimize directly with respect to $V$ (in the same way theat we could earlier for computing the norm itself); we temporarily assume that $U \in \rb^{N \times M}$ is known; we have:
   \BEAS
 & = & \min_{V   \in \rb^{P \times M} }  \frac{1}{2NP}\| Y - U V^\top \|_F^2 + \frac{\lambda}{2} \| V \|_F^2
\\
& = &     \frac{1}{2NP} \tr Y^\top \left[ \idm - 
 U ( U^\top U  + \lambda NP  \idm )^{-1} U^\top \right] Y  \\[-.1cm]
 \\
 & = &     \frac{1}{2NP}   \tr Y^\top  ( U U^\top/\lambda NP +   \idm )^{-1}   Y,
 \EEAS
 with a minimum attained at 
  $
  V = Y^\top U ( U^\top U + \lambda  NP \idm )^{-1}  = Y^\top  ( U U^\top + \lambda NP  \idm )^{-1}  U
  $.
The minimum is a \emph{convex} function of $UU^\top \in \rb^{N \times N}$ and we now have a convex optimization problem over \emph{positive semi-definite matrices}, which is equivalent to \eq{UVXinfinity}:
\BEQ
\label{eq:AA}
\min_{A \in \rb^{N\times N}, \ A \succcurlyeq 0 }
 \frac{1}{2NP} \tr Y^\top  ( A/\lambda NP+    \idm )^{-1}   Y +  \frac{ \lambda}{2} \min_{ A = \sum_{m \geqslant 0} u_m u_m^\top } \sum_{m\geqslant 0} F(u_m u_m^\top).
  \EEQ
It can be lower bounded by the following still convex, but now solvable in polynomial time, problem:
\BEQ
\label{eq:A} 
\min_{A \in \rb^{N\times N}, \ A \succcurlyeq 0}
 \frac{1}{2} \tr Y^\top  ( A/\lambda +    \idm )^{-1}   Y + \frac{ \lambda}{2} F(A).
  \EEQ
 This fully convex approach will be solved within a globally optimal low-rank optimization framework (presented in the next section). Then, rounding operations similar to \mysec{rounding} may be used to improve the solution---note that this rounding technique takes $Y$ into account and it thus preferable to the direct application of \mysec{rounding}.

\subsection{Low rank optimization over positive definite matrices}

\label{sec:lowrank}
We first smooth the problem by using $
  (1-\nu)   \sum_{i,j=1}^N (A_{ij}^2 + \varepsilon^2   )^{1/2} + \nu \tr A
$ as an approximation of $F(A)$,
and  $ (1-\nu)   ( \sum_{i=1}^N ( u_i^2 + \varepsilon^2 )^{1/2} )^2 + \nu \|u\|_2^2$ as an approximation of $F(uu^\top)$.

Following \cite{burer1}, since we expect low-rank solutions, we can optimize over low-rank matrices. Indeed, \cite{burer1} shows that if $G$ is a convex function over positive semidefinite symmetric matrices of size $N$, with a rank deficient global minimizer (i.e., of rank $r<N$), then the function $U \mapsto G(UU^\top)$ defined over matrices $U \in \rb^{N \times M}$ has no local minima as soon as $M>r$. The following novel proposition goes a step further for twice differentiable functions by showing that there is no need to know $r$ in advance:

\vspace*{-.1cm}

\begin{proposition}
\label{prop:conv_lowrank}
Let $G$ be a twice differentiable convex function  over positive semidefinite symmetric matrices of size $N$, with compact level sets.   If the function $H: U \mapsto G(UU^\top)$ defined over matrices $U \in \rb^{N \times M}$  has a local minimum at a \emph{rank-deficient} matrix $U$, then $U U^\top$ is a global minimum of~$G$.
\end{proposition}

\vspace*{-.1cm}

\begin{proof}
Let   $N = UU^\top$.
The gradient of $H$ is equal to $ \nabla H(U) = 2 \nabla G(UU^\top) U $ and the Hessian of $H$ is such that
$\nabla^2 H(U)(V,V) = 2 \tr \nabla G(UU^\top) VV^\top + \nabla^2 G(UU^\top) ( U V^\top + VU^\top, U V^\top + VU^\top)$. Since we have a local mimimum, 
 $ \nabla H (U) = 0 $  which implies that $\tr \nabla G(N) N  =  \tr \nabla H(U) U^\top = 0$.
Moreover, by invariance by post-multiplying $U$ by an orthogonal matrix,  without loss of generality, we may consider that the last column of $U$ is zero. We now consider all directions $V \in \rb^{ N \times M}$ with first $M-1$ columns equal to zero and last column being equal
to a given $v \in \rb^N$. The second order Taylor expansion of $H(U+tV)$ is
\BEAS
H(U+tV) 
&\!\! = \!\!&    H(U) +  t^2 \tr \nabla G(N) VV^\top   \\
& = & + \frac{t^2}{2} \nabla^2 G(N) ( U V^\top + VU^\top, U V^\top + VU^\top) + O(t^3)  \\[-.1cm]
& \!\!=\!\! &    H(U)  + t^2 v^\top \nabla G(N) v  + O(t^3). \EEAS
Since we have a local minima, we must have $v^\top \nabla G(N) v  \geqslant 0$. Since $v$ is arbitrary, this implies that $\nabla G(N) \succcurlyeq 0$. Together with the convexity of $G$ and $\tr \nabla G(N) N=0$, this implies that we have a global minimum of $G$~\cite{boyd}.
\end{proof}
The last proposition suggests to try a small $M$, and to check that a local minimum that we can obtain with descent algorithms is indeed rank-deficient. If it is, we have a solution; if not, we simply increase $M$ and start again until $M$ turns out to be greater than $r$.

Note that minimizing our convex lower bound in \eq{UVconvex} by any descent algorithm in $(U,V)$ is different than solving  directly \eq{UV}: in the first situation, there are no (non-global) local minima, whereas there may be some in the second situation.
In practice, we use a quasi-Newton algorithm which has complexity $O(N^2)$   to reach a stationary point, but requires to compute the Hessian of size $NM\times NM$ to check and potentially escape local minima.

 \subsection{Links with sparse principal component analysis}

 If we now consider that we want sparse dictionary elements instead of sparse decompositions, we exactly obtain the problem of sparse PCA~\cite{dspca,hastieSPCA}, where one wishes to decompose a data matrix $Y$ into $X=UV^\top$ where the dictionary elements are sparse, and thus easier to interpret. Note that in our situation, we have seen that with $\| \cdot\|_R = \| \cdot \|_2$, the problem in \eq{UV}  is equivalent to \eq{AA} and indeed only depends on the covariance matrix $\frac{1}{P} YY^\top$.
 
 This approach to sparse PCA is similar to the non convex formulations of~\cite{hastieSPCA} and is to be contrasted with the convex formulation of~\cite{dspca} as we aim at directly obtaining a \emph{full} decomposition of $Y$ with an implicit trade-off between  dictionary size (here the number of principal components) and sparsity of such components. Most works consider one unique component, even though the underlying data have many more underlying dimensions, and deal with multiple components by iteratively solving a reduced problem. In the non-sparse case, the two approaches are equivalent, but they are not here. By varying $\lambda$ and $\nu$, we obtain a set of solutions with varying ranks and sparsities. We are currently comparing the approach of   \cite{hastieSPCA}, which constrains the rank of the decomposition to ours, where the rank is penalized implicitly.

 \begin{table}
 \begin{center}
 \hspace*{-1cm}
 \begin{tabular}{|c|ccc|ccc|ccc|}
 \hline 
 \multicolumn{4}{|c}{ } & \multicolumn{3}{|c|}{$N=100$ } & \multicolumn{3}{c|}{ $N=200$}  \\
 \hline
 $\#$  & $P$ & $\!\!$ $M$ $\!\!$ & $S$     & \textsc{NoConv} & $\!\!$ \textsc{Conv-R} $\!\!$ & \textsc{Conv} & \textsc{NoConv} & $\!\!$ \textsc{Conv-R}  $\!\!$& \textsc{Conv}   \\ 
 \hline 
  $\!\!$ 1  $\!\!$ &  $\!\!$ 10  $\!\!$ &  $\!\!$ 10  $\!\!$ &  $\!\!$ 2  $\!\!$    & \textbf{ $\!\!\!$  -16.4 $\!\pm\!$ 5.7 $\!\!\!$  } & $\!\!\!$ -9.0 $\!\pm\!$ 1.9 $\!\!\!$   & $\!\!\!$ -6.5 $\!\pm\!$ 2.3 $\!\!\!$   & \textbf{ $\!\!\!$  -19.8 $\!\pm\!$ 2.3 $\!\!\!$  } & $\!\!\!$ -10.2 $\!\pm\!$ 1.6 $\!\!\!$   & $\!\!\!$ -7.1 $\!\pm\!$ 2.0 $\!\!\!$   \\ 
   $\!\!$ 2  $\!\!$ &  $\!\!$ 20  $\!\!$ &  $\!\!$ 10  $\!\!$ &  $\!\!$ 2  $\!\!$    & \textbf{ $\!\!\!$  -40.8 $\!\pm\!$ 4.2 $\!\!\!$  } & $\!\!\!$ -11.6 $\!\pm\!$ 2.6 $\!\!\!$   & $\!\!\!$ -5.6 $\!\pm\!$ 3.2 $\!\!\!$   & \textbf{ $\!\!\!$  -45.5 $\!\pm\!$ 2.0 $\!\!\!$  } & $\!\!\!$ -16.4 $\!\pm\!$ 1.4 $\!\!\!$   & $\!\!\!$ -7.0 $\!\pm\!$ 1.3 $\!\!\!$   \\ 
   $\!\!$ 3  $\!\!$ &  $\!\!$ 10  $\!\!$ &  $\!\!$ 20  $\!\!$ &  $\!\!$ 2  $\!\!$    & $\!\!\!$ -8.6 $\!\pm\!$ 3.6 $\!\!\!$   & \textbf{ $\!\!\!$  -9.0 $\!\pm\!$ 1.8 $\!\!\!$  } & $\!\!\!$ -8.4 $\!\pm\!$ 1.9 $\!\!\!$   & \textbf{ $\!\!\!$  -15.0 $\!\pm\!$ 2.7 $\!\!\!$  } & $\!\!\!$ -11.5 $\!\pm\!$ 1.5 $\!\!\!$   & $\!\!\!$ -10.5 $\!\pm\!$ 1.5 $\!\!\!$   \\ 
   $\!\!$ 4  $\!\!$ &  $\!\!$ 20  $\!\!$ &  $\!\!$ 20  $\!\!$ &  $\!\!$ 2  $\!\!$    & \textbf{ $\!\!\!$  -24.9 $\!\pm\!$ 3.3 $\!\!\!$  } & $\!\!\!$ -13.0 $\!\pm\!$ 0.7 $\!\!\!$   & $\!\!\!$ -10.4 $\!\pm\!$ 1.1 $\!\!\!$   & \textbf{ $\!\!\!$  -40.9 $\!\pm\!$ 2.2 $\!\!\!$  } & $\!\!\!$ -18.9 $\!\pm\!$ 0.8 $\!\!\!$   & $\!\!\!$ -14.8 $\!\pm\!$ 0.7 $\!\!\!$   \\ 
   $\!\!$ 5  $\!\!$ &  $\!\!$ 10  $\!\!$ &  $\!\!$ 40  $\!\!$ &  $\!\!$ 2  $\!\!$    & $\!\!\!$ -6.6 $\!\pm\!$ 2.8 $\!\!\!$   & $\!\!\!$ -8.9 $\!\pm\!$ 1.5 $\!\!\!$   & \textbf{ $\!\!\!$  -9.0 $\!\pm\!$ 1.4 $\!\!\!$  } & $\!\!\!$ -7.6 $\!\pm\!$ 2.6 $\!\!\!$   & \textbf{ $\!\!\!$  -10.1 $\!\pm\!$ 1.6 $\!\!\!$  } & $\!\!\!$ -9.9 $\!\pm\!$ 1.6 $\!\!\!$   \\ 
   $\!\!$ 6  $\!\!$ &  $\!\!$ 20  $\!\!$ &  $\!\!$ 40  $\!\!$ &  $\!\!$ 2  $\!\!$    & \textbf{ $\!\!\!$  -13.2 $\!\pm\!$ 2.6 $\!\!\!$  } & $\!\!\!$ -12.3 $\!\pm\!$ 1.4 $\!\!\!$   & $\!\!\!$ -11.5 $\!\pm\!$ 1.3 $\!\!\!$   & \textbf{ $\!\!\!$  -25.4 $\!\pm\!$ 3.0 $\!\!\!$  } & $\!\!\!$ -16.7 $\!\pm\!$ 1.3 $\!\!\!$   & $\!\!\!$ -15.6 $\!\pm\!$ 1.4 $\!\!\!$   \\ 
\hline 
   $\!\!$ 7  $\!\!$ &  $\!\!$ 10  $\!\!$ &  $\!\!$ 10  $\!\!$ &  $\!\!$ 4  $\!\!$    & $\!\!\!$ 1.7 $\!\pm\!$ 3.9 $\!\!\!$   & \textbf{ $\!\!\!$  -1.5 $\!\pm\!$ 0.5 $\!\!\!$  } & $\!\!\!$ -0.2 $\!\pm\!$ 0.2 $\!\!\!$   & \textbf{ $\!\!\!$  -1.9 $\!\pm\!$ 2.5 $\!\!\!$  } & $\!\!\!$ -1.7 $\!\pm\!$ 0.6 $\!\!\!$   & $\!\!\!$ -0.1 $\!\pm\!$ 0.1 $\!\!\!$   \\ 
   $\!\!$ 8  $\!\!$ &  $\!\!$ 20  $\!\!$ &  $\!\!$ 10  $\!\!$ &  $\!\!$ 4  $\!\!$    & \textbf{ $\!\!\!$  -16.7 $\!\pm\!$ 5.9 $\!\!\!$  } & $\!\!\!$ -1.4 $\!\pm\!$ 0.8 $\!\!\!$   & $\!\!\!$ -0.0 $\!\pm\!$ 0.0 $\!\!\!$   & \textbf{ $\!\!\!$  -27.1 $\!\pm\!$ 1.8 $\!\!\!$  } & $\!\!\!$ -3.0 $\!\pm\!$ 0.7 $\!\!\!$   & $\!\!\!$ 0.0 $\!\pm\!$ 0.0 $\!\!\!$   \\ 
   $\!\!$ 9  $\!\!$ &  $\!\!$ 10  $\!\!$ &  $\!\!$ 20  $\!\!$ &  $\!\!$ 4  $\!\!$    & $\!\!\!$ 2.2 $\!\pm\!$ 2.4 $\!\!\!$   & \textbf{ $\!\!\!$  -2.5 $\!\pm\!$ 0.9 $\!\!\!$  } & $\!\!\!$ -1.7 $\!\pm\!$ 0.8 $\!\!\!$   & $\!\!\!$ 2.0 $\!\pm\!$ 2.9 $\!\!\!$   & \textbf{ $\!\!\!$  -2.5 $\!\pm\!$ 0.8 $\!\!\!$  } & $\!\!\!$ -1.2 $\!\pm\!$ 1.0 $\!\!\!$   \\ 
   $\!\!$ 10  $\!\!$ &  $\!\!$ 20  $\!\!$ &  $\!\!$ 20  $\!\!$ &  $\!\!$ 4  $\!\!$    & $\!\!\!$ -1.2 $\!\pm\!$ 2.5 $\!\!\!$   & \textbf{ $\!\!\!$  -3.1 $\!\pm\!$ 1.1 $\!\!\!$  } & $\!\!\!$ -0.9 $\!\pm\!$ 0.9 $\!\!\!$   & \textbf{ $\!\!\!$  -12.1 $\!\pm\!$ 3.0 $\!\!\!$  } & $\!\!\!$ -5.5 $\!\pm\!$ 1.0 $\!\!\!$   & $\!\!\!$ -1.6 $\!\pm\!$ 1.0 $\!\!\!$   \\ 
   $\!\!$ 11  $\!\!$ &  $\!\!$ 10  $\!\!$ &  $\!\!$ 40  $\!\!$ &  $\!\!$ 4  $\!\!$    & $\!\!\!$ 3.5 $\!\pm\!$ 3.0 $\!\!\!$   & $\!\!\!$ -3.3 $\!\pm\!$ 1.3 $\!\!\!$   & \textbf{ $\!\!\!$  -3.3 $\!\pm\!$ 1.5 $\!\!\!$  } & $\!\!\!$ 2.6 $\!\pm\!$ 0.9 $\!\!\!$   & \textbf{ $\!\!\!$  -3.3 $\!\pm\!$ 0.5 $\!\!\!$  } & $\!\!\!$ -3.3 $\!\pm\!$ 0.5 $\!\!\!$   \\ 
   $\!\!$ 12  $\!\!$ &  $\!\!$ 20  $\!\!$ &  $\!\!$ 40  $\!\!$ &  $\!\!$ 4  $\!\!$    & $\!\!\!$ 3.7 $\!\pm\!$ 2.3 $\!\!\!$   & \textbf{ $\!\!\!$  -3.9 $\!\pm\!$ 0.6 $\!\!\!$  } & $\!\!\!$ -3.6 $\!\pm\!$ 0.8 $\!\!\!$   & $\!\!\!$ -1.7 $\!\pm\!$ 1.7 $\!\!\!$   & \textbf{ $\!\!\!$  -6.3 $\!\pm\!$ 0.9 $\!\!\!$  } & $\!\!\!$ -5.3 $\!\pm\!$ 0.8 $\!\!\!$   \\ 
 \hline 
  $\!\!$ 13  $\!\!$ &  $\!\!$ 10  $\!\!$ &  $\!\!$ 10  $\!\!$ &  $\!\!$ 8  $\!\!$    & $\!\!\!$ 9.6 $\!\pm\!$ 3.4 $\!\!\!$   & \textbf{ $\!\!\!$  -0.1 $\!\pm\!$ 0.1 $\!\!\!$  } & $\!\!\!$ 0.0 $\!\pm\!$ 0.0 $\!\!\!$   & $\!\!\!$ 7.2 $\!\pm\!$ 3.0 $\!\!\!$   & \textbf{ $\!\!\!$  -0.1 $\!\pm\!$ 0.1 $\!\!\!$  } & $\!\!\!$ 0.0 $\!\pm\!$ 0.0 $\!\!\!$   \\ 
   $\!\!$ 14  $\!\!$ &  $\!\!$ 20  $\!\!$ &  $\!\!$ 10  $\!\!$ &  $\!\!$ 8  $\!\!$    & \textbf{ $\!\!\!$  -1.6 $\!\pm\!$ 3.7 $\!\!\!$  } & $\!\!\!$ 0.0 $\!\pm\!$ 0.0 $\!\!\!$   & $\!\!\!$ 0.0 $\!\pm\!$ 0.0 $\!\!\!$   & \textbf{ $\!\!\!$  -4.8 $\!\pm\!$ 2.3 $\!\!\!$  } & $\!\!\!$ 0.0 $\!\pm\!$ 0.0 $\!\!\!$   & $\!\!\!$ 0.0 $\!\pm\!$ 0.0 $\!\!\!$   \\ 
   $\!\!$ 15  $\!\!$ &  $\!\!$ 10  $\!\!$ &  $\!\!$ 20  $\!\!$ &  $\!\!$ 8  $\!\!$    & $\!\!\!$ 9.6 $\!\pm\!$ 2.4 $\!\!\!$   & \textbf{ $\!\!\!$  -0.4 $\!\pm\!$ 0.4 $\!\!\!$  } & $\!\!\!$ -0.2 $\!\pm\!$ 0.3 $\!\!\!$   & $\!\!\!$ 9.4 $\!\pm\!$ 1.5 $\!\!\!$   & \textbf{ $\!\!\!$  -0.4 $\!\pm\!$ 0.4 $\!\!\!$  } & $\!\!\!$ -0.2 $\!\pm\!$ 0.2 $\!\!\!$   \\ 
   $\!\!$ 16  $\!\!$ &  $\!\!$ 20  $\!\!$ &  $\!\!$ 20  $\!\!$ &  $\!\!$ 8  $\!\!$    & $\!\!\!$ 11.3 $\!\pm\!$ 1.8 $\!\!\!$   & \textbf{ $\!\!\!$  -0.2 $\!\pm\!$ 0.2 $\!\!\!$  } & $\!\!\!$ -0.0 $\!\pm\!$ 0.0 $\!\!\!$   & $\!\!\!$ 7.0 $\!\pm\!$ 2.5 $\!\!\!$   & \textbf{ $\!\!\!$  -0.4 $\!\pm\!$ 0.3 $\!\!\!$  } & $\!\!\!$ -0.0 $\!\pm\!$ 0.0 $\!\!\!$   \\ 
   $\!\!$ 17  $\!\!$ &  $\!\!$ 10  $\!\!$ &  $\!\!$ 40  $\!\!$ &  $\!\!$ 8  $\!\!$    & $\!\!\!$ 8.8 $\!\pm\!$ 3.0 $\!\!\!$   & \textbf{ $\!\!\!$  -0.8 $\!\pm\!$ 0.7 $\!\!\!$  } & $\!\!\!$ -0.7 $\!\pm\!$ 0.7 $\!\!\!$   & $\!\!\!$ 7.2 $\!\pm\!$ 1.3 $\!\!\!$   & \textbf{ $\!\!\!$  -0.7 $\!\pm\!$ 0.4 $\!\!\!$  } & $\!\!\!$ -0.5 $\!\pm\!$ 0.5 $\!\!\!$   \\ 
   $\!\!$ 18  $\!\!$ &  $\!\!$ 20  $\!\!$ &  $\!\!$ 40  $\!\!$ &  $\!\!$ 8  $\!\!$    & $\!\!\!$ 10.9 $\!\pm\!$ 1.1 $\!\!\!$   & \textbf{ $\!\!\!$  -0.9 $\!\pm\!$ 0.6 $\!\!\!$  } & $\!\!\!$ -0.6 $\!\pm\!$ 0.5 $\!\!\!$   & $\!\!\!$ 9.4 $\!\pm\!$ 1.0 $\!\!\!$   & \textbf{ $\!\!\!$  -1.0 $\!\pm\!$ 0.4 $\!\!\!$  } & $\!\!\!$ -0.4 $\!\pm\!$ 0.4 $\!\!\!$   \\ 
 \hline 
  \end{tabular}
 \end{center}
 
 \vspace*{-.3cm}
 
 \caption{Percentage of improvement in mean squared error, with respect to spectral denoising, for various parameters of the data generating process. See text
 for details.}
 \label{tab:results}
 \end{table}

\section{Simulations}
 \label{sec:simulations}

 We have performed extensive simulations on synthetic examples to compare the various formulations.  Because of identifiability problems which are the subject of ongoing work, it is not appropriate to compare decomposition coefficients and/or dictionary elements; we rather consider a denoising experiment. Namely, we have generated matrices $Y_0=UV^\top$ as follows: select $M$ unit norm dictionary elements $v_1,\dots,v_M$ in $\rb^P$ uniformly and independently at random, for each $n \in \{1,\dots,N\}$, select $S$ indices in $\{1,\dots,M\}$ uniformly at random and form the $n$-th row of $U \in \rb^{N \times M}$ with zeroes except for random normally distributed elements at the $S$ selected indices. Construct $Y = Y_0 +{(\tr Y_0 Y_0^\top)^{1/2}} \sigma \varepsilon / {(NP)^{1/2}}$, where $\varepsilon$ has independent standard normally distributed elements and $\sigma$ (held fixed at $0.6$). The goal is to estimate $Y_0$ from $Y$, and we compare the three following formulations on this task: (a) the convex minimization of \eq{A} through techniques presented in \mysec{lowrank} with varying $\nu$ and $\lambda$, denoted as \textsc{Conv}, (b) the rounding of the previous solution using techniques described in \mysec{rounding}, denoted as \textsc{Conv-R}, and (c) the low-rank constrained problem in \eq{UV} with $\| \cdot \|_C = \| \cdot \|_1$ and
 $\| \cdot \|_R = \| \cdot \|_2$ with varying $\lambda$ and $M$, denoted as \textsc{NoConv}, and which is the standard method in sparse dictionary learning~\cite{olshausen,elad,ng}.
 
 For the three methods and for each replication, we select the two regularization parameters that lead to the minimum value $\| X - Y_0 \|^2$, and compute the relative improvement on using the singular value decomposition (SVD) of $Y$. If the value is negative, denoising is better than with the SVD (the more negative, the better). In Table~\ref{tab:results}, we present averages over 10 replications for various values of $N$, $P$, $M$, and $S$.
 
First, in these simulations where the decomposition coefficients are known to be sparse, penalizing by $\ell^1$-norms indeed improves performance on spectral denoising for all methods.
Second, as expected, the rounded formulation (\textsc{Conv-R})  does perform better than the non-rounded one   (\textsc{Conv}), i.e., our rounding procedure allows to find ``good'' local minima of the non-convex problem in \eq{UV}.

 Moreover, in   high-sparsity situations ($S=2$, lines 1 to 6 of Table~\ref{tab:results}), we see that the rank-constrained formulation \textsc{NoConv} outperforms the convex formulations, sometimes  by a wide margin (e.g., lines 1 and 2). This is not the case when the ratio $M/P$ becomes larger than 2 (lines 3 and 5). In the medium-sparsity situation ($S=4$, lines 7  to 12), we observe the same phenomenon, but the non-convex approach is better only when the ratio $M/P$ is smaller than or equal to one. Finally, in low-sparsity situations ($S=8$, lines 13 to 18), imposing sparsity does not improve performance much and the local minima of the non-convex approach $\textsc{NoConv}$ really hurt performance.
 Thus, from Table~\ref{tab:results}, we can see that with high sparsity (small $S$) and small relative dictionary size of the original non noisy data (i.e., low ratio $M/P$), the non convex approach performs better. We are currently investigating theoretical arguments to support these empirical findings.

\section{Conclusion}

In this paper, we have investigated the possibility of convexifying the sparse dictionary learning problem. We have reached both positive and negative conclusions: indeed, it is possible to convexify the problem by letting the dictionary size  explicitly grow with proper regularization to ensure low rank solutions; however, it only leads to better predictive performance for problems which are not too sparse and with large enough dictionaries. In the high-sparsity/small-dictionary cases, the non convex problem is empirically simple enough to solve so that our convexification leads to no gain.

We are currently investigating more refined convexifications and extensions to nonnegative variants~\cite{ding}, applications of our new decomposition norms to clustering~\cite{ding}, the possibility of obtaining consistency theorems similar to \cite{tracenorm} for the convex formulation, and the application to the image denoising problem~\cite{elad}.

 
\bibliography{matrix_factorization}

\end{document}